\documentclass{article} % For LaTeX2e
\usepackage{nips14submit_e,times}
\usepackage{hyperref}
\usepackage{url}

\usepackage{amssymb,amsmath,amsthm}

\usepackage[ruled,vlined,linesnumbered]{algorithm2e}
\usepackage[T1]{fontenc}
\usepackage{graphicx}
\usepackage{graphics}
\usepackage{multirow}
\usepackage{fancyhdr}
\usepackage{bm}
\usepackage{subfigure}

\usepackage{pgfplots}
\usepackage{etex}
\reserveinserts{18}
\usepackage{morefloats}
\usepackage{array}
\usepackage{colortbl}

\usepackage{atveryend}
\makeatletter
\let\origcitation\citation
\AtEndDocument{\def\mycites{}%
  \def\citation#1{\g@addto@macro\mycites{#1^^J}\origcitation{#1}}}
\AtVeryEndDocument{\newwrite\citeout\immediate\openout\citeout=\jobname.cit
  \immediate\write\citeout{\mycites}\immediate\closeout\citeout}
\makeatother
\usepackage{tikz}
\usepackage{tikz-3dplot}
\usetikzlibrary{fadings}

\tikzfading[name=radialfade, inner color=transparent!0, outer color=transparent!100]

\DeclareMathOperator*{\argmin}{argmin}

\newcommand{\cO}{\ensuremath{\mathcal{O}}}
\newcommand{\cI}{\ensuremath{\mathcal{I}}}

\newcommand{\R}{\ensuremath{\mathbb{R}}}

\newcommand\T{\rule{0pt}{2.6ex}}
\newcommand\B{\rule[-1.2ex]{0pt}{0pt}}
\newcommand{\quotes}[1]{``#1''}

\newtheorem{theorem}{Theorem}

\theoremstyle{definition}

\title{Complexity Issues and Randomization Strategies in Frank-Wolfe Algorithms for Machine Learning}

\author{
Emanuele Frandi\\
Department of Electrical Engineering (ESAT-STADIUS)\\
KU Leuven, Belgium\\
\texttt{emanuele.frandi@esat.kuleuven.be}
\And
Ricardo \~Nanculef\\
Departamento de Inform\'atica\\
Universidad T\'ecnica Federico Santa Mar\'ia, Chile \\
\texttt{jnancu@inf.utfsm.cl}
\And
Johan Suykens\\
Department of Electrical Engineering (ESAT-STADIUS)\\
KU Leuven, Belgium\\
\texttt{johan.suykens@esat.kuleuven.be}
}

\nipsfinalcopy % Uncomment for camera-ready version

\begin{document}

\maketitle

\begin{abstract}
Frank-Wolfe algorithms for convex minimization have recently gained considerable attention from the Optimization and Machine Learning communities, as their properties make them a suitable choice in a variety of applications. 
However, as each iteration requires to optimize a linear model, a clever implementation is crucial to make such algorithms viable on large-scale datasets. For this purpose, approximation strategies based on a random sampling have been proposed by several researchers. In this work, we perform an experimental study on the effectiveness of these techniques, analyze possible alternatives and provide some guidelines based on our results.
\end{abstract}

\section{Introduction}

The Frank-Wolfe algorithm \cite{wolfe1954}, hereafter denoted as FW, is a general method to solve
\[
\min_{\alpha \in \Sigma} \; f(\alpha),
\]
where $f: \R^m \rightarrow \R$ is a convex differentiable function, and $\Sigma \subset \R^m$ is a convex polytope. 
Given the current iterate ${\alpha}^{(k)} \in \Sigma$, a standard {FW} iteration consists of the following steps: 
%\pause
\vskip 0.1cm
%\end{itemize}
\begin{enumerate}		
\item Define a search direction $d^{(k)}$ by optimizing a linear model: 
\begin{equation}\label{linear_subpb}
u^{(k)} %= \argmin_{u \in \Sigma} f({\alpha}_k) + \nabla f({\alpha}_k)^T({u} - {\alpha}_k) 
\in \argmin_{u \,\in\, \Sigma} \, (u-\alpha^{(k)})^T \nabla f({\alpha}^{(k)}) = \argmin_{u \, \in \, \mathcal{V}(\Sigma)} u^{T} \nabla f(\alpha^{(k)}), \;\; {d}^{(k)} = {u}^{(k)} - {\alpha}^{(k)}, 
\end{equation}
where $\mathcal{V}(\Sigma)$ denotes the set of vertices of $\Sigma$.
\item Choose a stepsize $\lambda^{(k)}$, e.g. by a line-search: $\lambda^{(k)} \in  \argmin_{\lambda \,\in\, [0,1]}  f({\alpha}^{(k)} + \lambda
{d}^{(k)})$. %(note: it's actually optional, we can just choose $\lambda = \frac{2}{2+k}$).
\item Update: 
%\[
${\alpha}^{(k+1)}=  {\alpha}^{(k)} + \lambda^{(k)} {d}^{(k)} = (1-\lambda^{(k)}){\alpha}^{(k)} + \lambda^{(k)} {u}^{(k)} \,.$
%\]
\end{enumerate}

Recently, the Optimization and Machine Learning communities have showed a renewed surge of interest in the family of FW algorithms \cite{Jaggi2013ICMLa,harchaoui14,SWAP_paper}. They enjoy bounds on the number of iterations which are independent of the problem size, as well as sparsity guarantees \cite{clarkson_coresets,Jaggi2013ICMLa}. Furthermore, variants of the above basic procedure exist which attain a linear convergence rate \cite{wolfe1970,GuelatMarcotte,SWAP_paper,jaggi_linconv}.
% It's slow, but enjoys solid convergence properties, further enhanced by variants. It also enforces sparsity. The number of its has a bound independent from the pb size. 
Such properties make FW a good choice for problems arising in a variety of applications \cite{signoretto14book,IJPRAI11,Jaggi2013ICMLb}. 

\textbf{Complexity of Frank-Wolfe Iterations.}
%A standard FW iteration requires the optimization of a linear model. 
As the total number of FW iterations can be large in practice, devising a convenient way to find a solution to the subproblem (\ref{linear_subpb}) is often mandatory in order to make the algorithm viable. A typical situation arises when (\ref{linear_subpb}) has an analytical solution or the problem structure makes it easy to solve \cite{SWAP_paper, rinaldi13}. Still, the resulting complexity can be impractical when handling large-scale data. As a motivating example, we consider the problem %quadratic convex problem
\begin{equation}\label{svm_problem}
\min_{\alpha \,\in\, \R^{m}} \;\; f(\alpha) = \tfrac{1}{2}  \alpha^{T} K \alpha\ \;\;\;\; \mbox{s.t. } \sum_{i=1}^m \alpha_i = 1,
\, \; \alpha \geq 0 \; ,
\end{equation}
which stems from the task of training a nonlinear $L_2$-SVM model for binary classification \cite{coreSVMs05tsang,CIARP}. Here, $K$ is a positive definite kernel matrix.
In this case, $\mathcal{V}(\Sigma) = \{e_{1},\ldots,e_{m}\}$, hence we have
\[
u^{(k)} = e_{i_{*}^{(k)}}, \qquad \mbox{where} \qquad i_{*}^{(k)} \in \argmin_{i=1,\ldots,m} \nabla f(\alpha^{(k)})_i = \argmin_{i=1,\ldots,m} \; \sum_{j \,|\, \alpha^{(k)}_j > 0} K_{i,j}\alpha^{(k)}_j \,.
\]
The theoretical cost of an iteration is therefore $\mathcal{O}(m|\mathcal{I}^{(k)}|)$, % \leq \mathcal{O}(mk)$, 
where $\mathcal{I}^{(k)} = \{i\, | \, \alpha^{(k)}_{i} > 0 \}$, proportional to the number of examples.\footnote{More in general, it is proportional to $|\mathcal{V}(\Sigma)|$ and to the cost of computing $u^{T}\nabla f(\alpha^{(k)})$, with $u \in \mathcal{V}(\Sigma)$.} In order to circumvent the dependence from the dataset size, the use of approximation strategies based on a random sampling has been proposed by several researchers \cite{coreSVMs05tsang,IJPRAI11}, but, up to our knowledge, never systematically studied on practical problems. We attempt to fill this gap by performing an experimental study on the effect of using such techniques.

\section{Randomization Strategies and Possible Alternatives}\label{sec2}

In this section, we consider two different techniques to reduce the computational effort in each FW iteration,
%\begin{enumerate}
%\item solving an approximate model by exploring only a fixed number of vertices of $\Sigma$;
%\item keeping the linear model updated in a smart way across iterations,
%\end{enumerate}
and try to identify the kind of problems where each can be applied effectively.

\subsection{Random Working Set Selection} A simple and yet effective way to avoid the dependence on $m$ is to explore only a fixed number of points in $\mathcal{V}(\Sigma)$. %\footnote{ccc}
In the case of (\ref{svm_problem}), this means extracting a sample $\mathcal{S} \subseteq \{1,\ldots,m\}$ and solving
\[
i^{(k)}_\mathcal{S} \in \argmin_{i \in \mathcal{S}} \nabla f(\alpha^{(k)})_i \,.
\]
The iteration cost becomes in this case $\mathcal{O}(|\mathcal{S}||\mathcal{I}^{(k)}|)$. % \leq \mathcal{O}(|S|k)$. 
The following result motivates this kind of approximation, suggesting that it is reasonable to keep the samples very small, i.e. to pick $|\mathcal{S}| \ll m$.
%, and that of the overall procedure by $\cO(|S|/\varepsilon^2)$.

\vspace{0.1cm}
\begin{theorem}[\cite{Smola01Learning}, Theorem 6.33]
%Let $D:=\{d_1,\ldots,d_m\} \subset \mathbb{R}$ be a set of
Let $\mathcal{D} \subset \mathbb{R}$ be a set of cardinality $m$, and let $\mathcal{D}^{\prime} \subset \mathcal{D}$ be a random subset
of size $r$. Then, the probability that the smallest element in $\mathcal{D}^{\prime}$ is less than or equal to $\tilde m$ elements of $\mathcal{D}$ is at least $1-(\frac{\tilde m}{m})^r$.
\end{theorem}
In the case of (\ref{svm_problem}), where
$\mathcal{D} = \{\nabla f(\alpha^{(k)})_1,\ldots,\nabla f(\alpha^{(k)})_m\}$ and $\mathcal{D}^{\prime} = \{ \nabla f(\alpha^{(k)})_i \, | \, i \in \mathcal{S} \}$, this means that, for example, it only takes $|\mathcal{S}| \approx 60$ to guarantee that, with probability at least $0.95$ (and independently of $m$), $\nabla f(\alpha^{(k)})_{i^{(k)}_\mathcal{S}}$ lies between the $5\%$ smallest gradient components. 

\textbf{Choice of the Stopping Criterion and Implications.} The stopping criterion for FW algorithms is usually based on the duality gap \cite{Jaggi2013ICMLa}:
%\begin{equation*}
%\Delta_d(\alpha^{(k)}) := (\alpha^{(k)})^{T}\nabla f(\alpha^{(k)}) - \min_{i} \nabla f(\alpha^{(k)})_i = -(d^{(k)})^T \nabla f(\alpha^{(k)})  \leq \varepsilon \ ,
%\end{equation*}
\begin{equation*}
\Delta_d(\alpha^{(k)}) := \max_{u \,\in \,\Sigma} \,(\alpha^{(k)}-u)^T \nabla f(\alpha^{(k)}) \stackrel{(\ref{svm_problem})}{=} 2 f(\alpha^{(k)}) - \nabla f(\alpha^{(k)})_{i_{*}^{(k)}}  \leq \varepsilon \ .
%(\alpha^{(k)})^{T}\nabla f(\alpha^{(k)}) - \nabla f(\alpha^{(k)})_{i_{*}^{(k)}}  \leq \varepsilon \ .
\end{equation*}
This criterion, however, is not applicable without computing the entire gradient $\nabla f(\alpha^{(k)})$, which is not done in the randomized case. As a possible alternative, we can use the approximate quantity
\[
\Delta_\mathcal{S}(\alpha^{(k)}) := 2 f(\alpha^{(k)}) - \nabla f(\alpha^{(k)})_{i^{(k)}_\mathcal{S}}.
%(\alpha^{(k)})^{T}\nabla f(\alpha^{(k)}) - \nabla f(\alpha^{(k)})_{i^{(k)}_\mathcal{S}}.
\]
Since $\Delta_\mathcal{S}(\alpha^{(k)}) \leq \Delta_d(\alpha^{(k)})$, this simplification entails a tradeoff between the reduction in computational cost and risk of an anticipated stopping. Although this can be considered acceptable in contexts such as SVM classification, where solving the optimization problem with a high accuracy is usually not needed, it is important to make sure that the impact of this approximation can be kept to an acceptable level. The experiments in the next section aim precisely at investigating this issue.

\subsection{Analytical Gradient Update}\label{sec2.1}
Another possibility to obtain a more efficient iteration is to exploit the structure of the problem to keep the exact gradient $\nabla f(\alpha^{(k)})$ updated at each iteration \cite{Kumar2011}. In the case of problem (\ref{svm_problem}), this can be done in $\cO(m)$ operations, since it is easy to see by using the formula for the FW step that
\[
\nabla f(\alpha^{(k+1)})_i = (1-\lambda^{(k)}) \nabla f(\alpha^{(k)})_i + \lambda^{(k)} K_{i,i^{(k)}_{*}}\,, \qquad i = 1,\ldots,m.
\]
Compared to a naive implementation, we get rid of a factor $|\cI^{(k)}|$ and, as an important by-product, we have that the duality gap can be updated exactly without any additional cost. %\cite{StochasticFW10}

%%Discutere del significato del dual gap in 2 righe? (vedi sotto)

%%Can't we use another stopping criterion? Yes, we can use a classical relative distance btw iterates or btw function values. However, we would not have a bound on the primal gap anymore, so it would not be easy to relate eps to the vicinity to the optimum or to the expected total number of iterations. Then of course one would object that even the approximate gap is not a valid bound for the primal gap (it can drop below 0, etc)...

\section{Numerical Results}

In order to assess the effectiveness of the above implementations of the FW step, we conducted numerical tests on the benchmark datasets \textbf{Adult a9a} ($m = 32561$), \textbf{Web w8a} ($m = 49749$), \textbf{IJCNN} ($m = 49990$) and \textbf{USPS-ext} ($m = 266079$) \cite{SVMLIB,UCI2010}. All the experiments were coded in C++, and executed on a 3.40GHz 4-core Intel machine with 16GB RAM running Linux.
%MENZIONARE GLI IPERPARAMETRI? O E' SPRECO DI SPAZIO?

Table \ref{exp_1} presents the statistics (averaged over $10$ runs) for classification accuracy on the test set, CPU time, number of iterations and support vectors, obtained with samplings of increasing size. The tolerance parameter was set to $\varepsilon = 10^{-4}$, and a Gaussian kernel was used for all the experiments. An LRU caching strategy was implemented to avoid the computation of recently used entries of $K$.

\begin{table}[h!!!]
\centering
\begin{small}
{\begin{tabular}{@{}lllllll}
\hline
\T {Dataset} & & $m$ points & 1000 points & 500 points & 250 points & 125 points \B\\
\hline
\textbf{Adult a9a} \T  & Test acc (\%) & $83.56$ & $84.10$ & $83.91$ & $83.68$ & $83.88$ \\
	 & Time (s) & $1.40e+02$ & $4.26e+02$  & $2.55e+02$  & $1.62e+02$ & $1.12e+02$ \\
  	& Iter & $2.02e+04$& $1.94e+04$ & $1.91e+04$ & $1.85e+04$ & $1.71e+04$ \\  
  	& SVs & $1.40e+04$& $1.37e+04$ & $1.36e+04$ & $1.34e+04$ & $1.20e+04$ \B \\
\hline
\textbf{Web w8a} \T  & Test acc (\%) & $99.36$ & $99.30$ & $99.28$ & $99.00$ & $98.49$ \\
	 & Time (s)  & $3.17e+02$  &  $2.50e+02$  & $1.60e+02$ & $5.55e+01$ & $2.75e+01$ \\
	& Iter & $1.65e+04$ & $1.39e+04$ & $1.24e+04$ & $4.63e+03$ & $2.17e+03$ \\  
  	& SVs & $6.43e+03$ & $6.33e+03$ & $5.77e+03$ & $2.82e+03$ & $1.70e+03$ \B \\
\hline
\textbf{IJCNN} \T & Test acc (\%) & $98.24$ & $98.42$ & $98.30$ & $98.28$ & $97.57$ \\
	 & Time (s)  &  $4.99e+01$ & $1.19e+02$ & $5.80e+01$ & $3.12e+01$ & $1.43e+01$  \\
  	& Iter & $1.61e+04$ & $1.46e+04$ & $1.22e+04$ & $9.91e+03$ & $5.69e+03$  \\  
  	& SVs & $3.17e+03$& $3.59e+03$ & $3.72e+03$ & $3.84e+03$  & $3.37e+03$ \B \\
\hline
\textbf{USPS-ext} \T & Test acc (\%) & $99.52$  & $98.90$ & $98.88$ & $99.50$ & $99.45$ \\
 	 & Time (s)  & $1.77e+03$ &  $4.25e+02$ &  $2.83e+02$ & $1.56e+02$ & $4.97e+01$  \\
  	& Iter & $2.05e+04$ & $9.07e+03$ & $4.32e+03$ & $2.70e+03$ & $1.65e+03$ \\  
  	& SVs & $3.94e+03$ & $3.59e+03$ & $3.00e+03$ & $2.37e+03$ & $1.60e+03$ \B \\
\hline
\end{tabular}}
\caption{\label{exp_1} Average statistics with different sampling sizes.}
\end{small}
\end{table}

%%Observations:
First of all, note that the effect of sampling is substantially problem-dependent. On some datasets, such as \textbf{USPS-ext}, FW clearly encounters an early stopping even with a fairly large sampling size, while other results, such as those on \textbf{Adult a9a}, appear more stable. In some cases, e.g. on \textbf{Web w8a}, there seems to be a cutoff point after which the performance degrades considerably. 
Still, some general trends can be estabilished: the number of iterations decreases monotonically with $|\mathcal{S}|$, as expected from the observations in Section \ref{sec2}, and CPU times decrease accordingly.
On the contrary, as seen from the results on \textbf{IJCNN}, the model size is not always monotonic with respect to $|\mathcal{S}|$. This arguably happens because solving (\ref{linear_subpb}) approximately can lead to spurious points being selected as FW vertices. Finally, note that the full sampling solution (which employs the strategy in Section \ref{sec2.1}) is very competitive on the smaller problems, while it is still very time consuming on the largest dataset \textbf{USPS-ext}. This intuitively suggests that a random sampling is computationally convenient when it can still produce a good solution with $|\mathcal{S}| \ll m / \mu_{|\mathcal{I}^{(k)}|}$, where $\mu_{|\mathcal{I}^{(k)}|}$ is an estimate of the average cardinality of $\cI^{(k)}$ across iterations. Some of these conclusions are summarized in Table \ref{recom}. 

\begin{figure}[h!!!]
\begin{center}
\subfigure[]{
{\includegraphics[scale = 0.099]{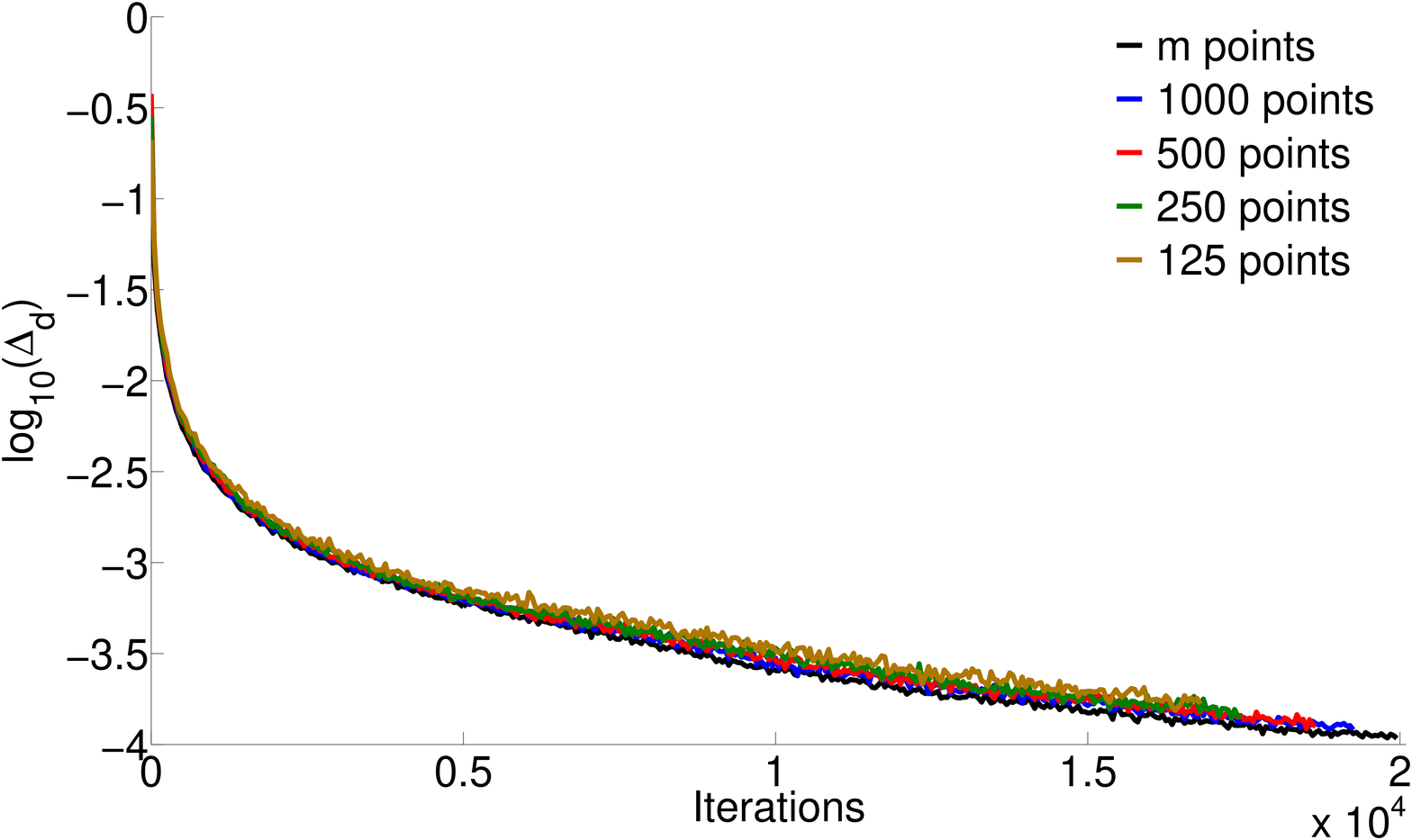}}}
\subfigure[]{
   {\includegraphics[scale = 0.099]{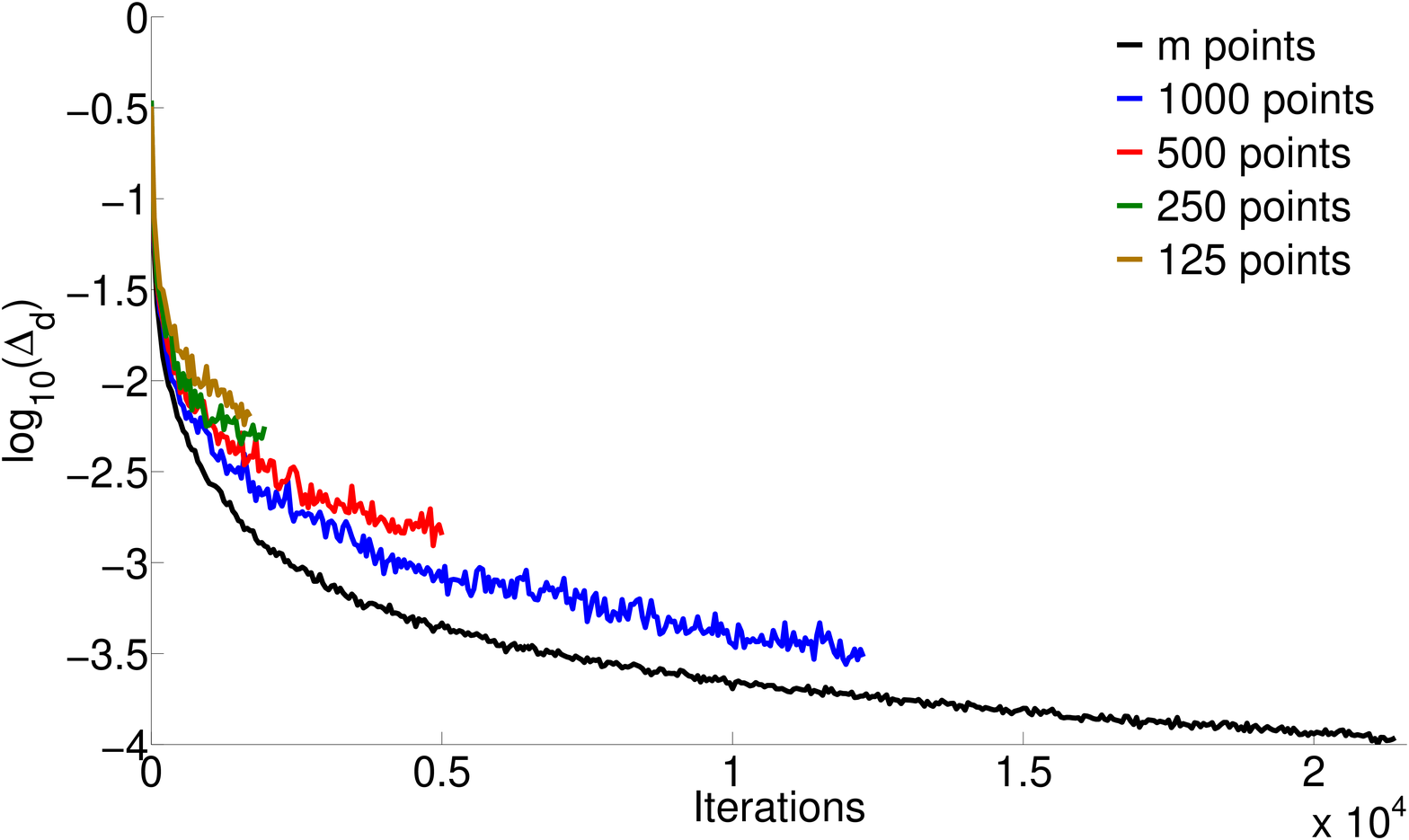}}}
\caption{\label{dual_path} Exact duality gap path on datasets \textbf{Adult a9a} (a) and \textbf{USPS-ext} (b).} 
\end{center}
\end{figure}

\begin{figure}[h!!!]
\begin{center}
\subfigure[]{
{\includegraphics[scale = 0.099]{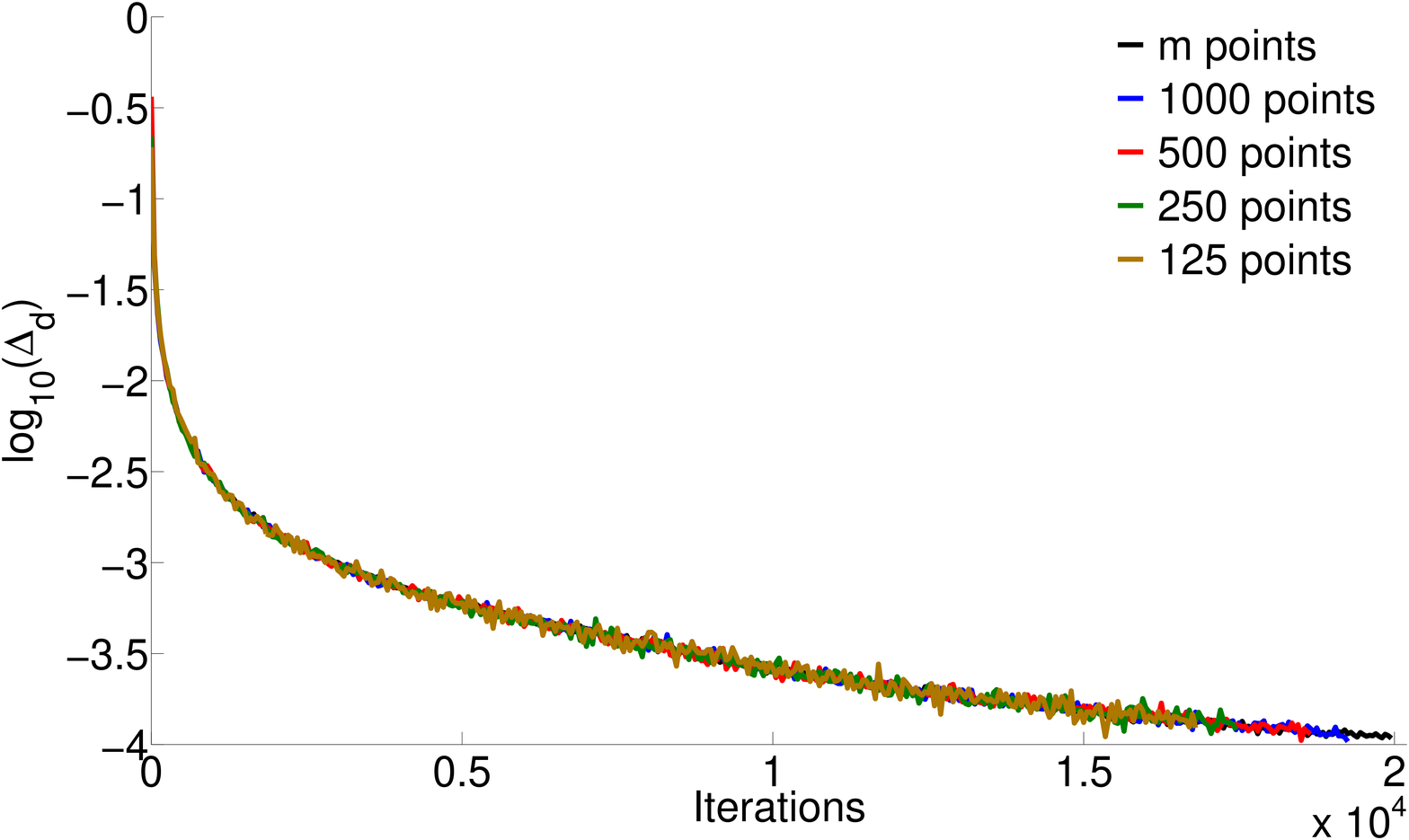}}}
\subfigure[]{
   {\includegraphics[scale = 0.099]{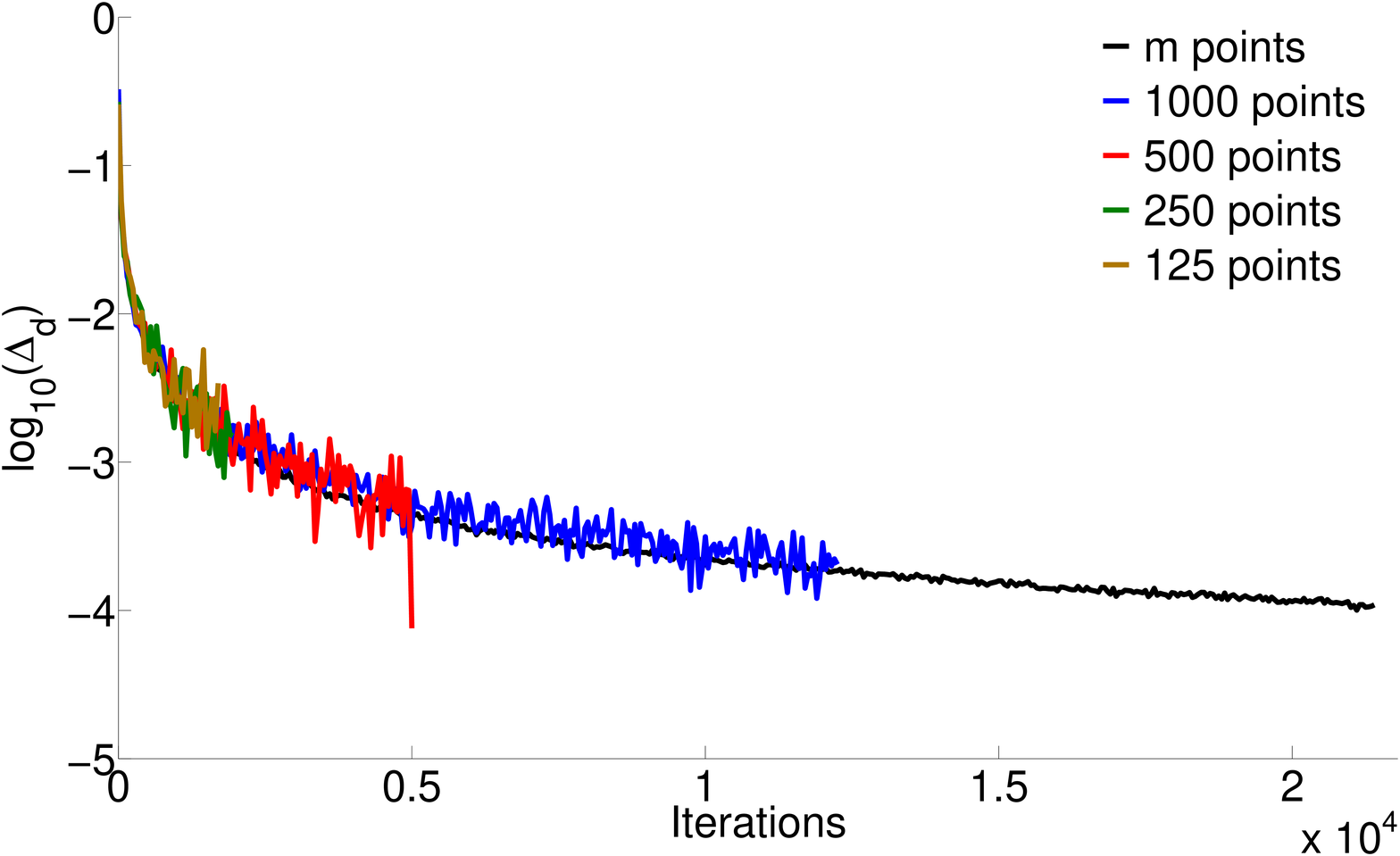}}}
\caption{\label{gradient_path} Approximate duality gap path on datasets \textbf{Adult a9a} (a) and \textbf{USPS-ext} (b).} 
\end{center}
\end{figure}

In the next experiment, we analyze, on the datasets \textbf{Adult a9a} and \textbf{USPS-ext}, the effect of sampling on the computation of the duality gap (and therefore on the stopping criterion) and on the minimization of the linear model. Figures \ref{dual_path} and \ref{gradient_path} report, respectively, the exact gap $\Delta_{d}$ and %the difference between $\nabla f(\alpha)_{i^*}$ and $\nabla f(\alpha)_{i_S}$
the approximate gap $\Delta_{\mathcal{S}}$, plotted in logarithmic scale against the iteration number for various sampling sizes.

The figures shed light on the results in Table \ref{exp_1}. On the dataset \textbf{Adult a9a}, the randomized strategy appears very effective: the duality gap does not deviate much from the ideal figure obtained with the full dataset, even for small sampling sizes. Furthermore, there are no significant differences between computing the exact and approximate duality gap. On the other hand, on \textbf{USPS-ext}, $\Delta_{d}$ is noticeably larger than its approximate counterpart, indicating that the algorithm is making less progress than predicted by $\Delta_{\mathcal{S}}$. Furthermore, the approximate gap exhibits large oscillations due to the random nature of the sampling, and it is possible that an \quotes{unlucky} iteration leads to a premature stopping, as can be seen from the figure. 
%NEW COMMENTS
It is interesting to note that the degradation in optimization quality (as measured by $\Delta_{d}$) is not reflected in this case by a corresponding loss in test accuracy, which is a phenomenon typical of classification problems. However, this is not true in general, as other applications such as function estimation are known to be more sensitive to a less accurate solution. 

\begin{table}[h!!!]
\centering
\begin{small}
{\begin{tabular}{@{}ll}
%\hline
%\T {Dataset} & & $m$ points & 1000 points & 500 points & 250 points & 125 points \B\\
\hline
\textbf{Randomized Working Set Selection} \T  & - Applicable whenever $\Sigma$ is a polytope \\
 & - Large computational gain when $|\mathcal{S}| \ll m / \mu_{|\mathcal{I}^{(k)}|}$ \\
 & - Performance depends on the problem \B \\
\hline
\textbf{Analytical Gradient Update} \T  & - Convenient for structured $f$ (e.g. quadratic) \\
 & -  Saves a factor  $|\cI^{(k)}|$ at each iteration \\
& - Deterministic results \B \\
\hline
\end{tabular}}
\caption{\label{recom} Some recommendations on the implementation of the FW step.}
\end{small}
\end{table}

\textbf{Adaptive Strategies.} Taking into account all the above, one would ideally want to be able to select an optimal strategy automatically, based on the data and the actual performance. Provided both strategies can be applied to the problem at hand, one could for example start by performing a fixed number $\bar k > 0$ of iterations using both, 
and then devise some criterion based on the difference in duality gap to decide whether the approximation is adequate. However, a discussion on how to effectively implement such a strategy would be nontrivial, and as such is deferred to a separate work.

% Non voglio mettere troppi dettagli perchŽ non sarei in grado di spiegarli con lo spazio a disposizione. Preferisco accennare in poche parole a quello che vogliamo fare e lasciare la curiositˆ per un lavoro successivo che provare a fare una proposta debole e poco ragionata in cinque righe.

\section{Conclusions}
Using SVM classification problems as a motivation, we have performed an experimental study on the effectiveness and impact of some techniques designed to alleviate the computational burden of the optimization step in a FW iteration. Our results suggested that, while it comes with some caveats, a random sampling technique may be the most viable choice on very large-scale problems. %, such as ... 
On the other hand, when the problem size  is not prohibitive (e.g. batch training tasks with medium to large datasets), fast updating schemes which exploit the problem structure might provide a better choice.

\section*{Acknowledgments}
The research leading to these results has received funding from the European Research Council under the European Union's Seventh Framework Programme (FP7/2007-2013) / ERC AdG A-DATADRIVE-B (290923). This paper reflects only the authors' views and the Union is not liable for any use that may be made of the contained information. Research Council KUL: GOA/10/09 MaNet, CoE PFV/10/002 (OPTEC),
BIL12/11T; Flemish Government: FWO: projects: G.0377.12 (Structured systems), G.088114N (Tensor based data similarity); PhD/Postdoc grants; iMinds Medical Information Technologies SBO 2014; IWT:  POM II SBO 100031; Belgian Federal Science Policy Office: IUAP P7/19 (DYSCO, Dynamical systems, control and optimization, 2012-2017).

% Aggiungere che l'update analitico del gradiente andrebbe in ogni caso implementato ovunque sia possibile (non ci sono svantaggi ad applicare questa tecnica,  solo una riscrittura ``furba'' del passo di ottimizzazione. Per˜ non sempre si pu˜ fare (per es. se la funzione  generica), mentre esplorare un numero di vertici random del poliedro ammissibile  sempre fattibile.

\begin{footnotesize}

\bibliographystyle{plain}	
\bibliography{NIPS14_bibliography}

\end{footnotesize}

\end{document}